\documentclass{article}
\usepackage{ijcai24}

\usepackage{times}
\usepackage{soul}
\usepackage{url}
\usepackage[hidelinks]{hyperref}
\usepackage[utf8]{inputenc}
\usepackage[small]{caption}
\usepackage{graphicx}
\usepackage{amsmath}
\usepackage{amsthm}
\usepackage{booktabs}
\usepackage{algorithm}
\usepackage{algorithmic}
\usepackage[switch]{lineno}

\usepackage{amsmath}
\usepackage{amssymb}
\usepackage{mathtools}
\usepackage{amsthm}

\DeclareMathOperator*{\argmax}{arg\,max}

\title{\textit{SQT} -- \textit{std} $Q$-target.}

\author{
    \affiliations
    \emails
}

\author{
Nitsan Soffair$^1$
\and
Dotan Di-Castro$^2$\and
Orly Avner$^{2}$\And
Shie Mannor$^1$\\
\affiliations
$^1$Technion\\
$^2$Bosch\\
\emails
Nitsan.Soffair@gmail.com,
\{Dotan.DiCastro, Orly.Avner\}@gmail.com,
Shie@EE.Technion.ac.il 
}

\begin{document}

\maketitle

\begin{abstract}
    \textit{Std} $Q$-target is a \textit{conservative}, actor-critic, ensemble, $Q$-learning-based algorithm, which is based on a single key $Q$-formula: $Q$-networks standard deviation, which is an "uncertainty penalty", and, serves as a minimalistic solution to the problem of \textit{overestimation} bias. We implement \textit{SQT} on top of TD3/TD7 code and test it against the state-of-the-art (SOTA) actor-critic algorithms, DDPG, TD3 and TD7 on seven popular MuJoCo and Bullet tasks. Our results demonstrate \textit{SQT}'s $Q$-target formula superiority over \textit{TD3}'s $Q$-target formula as a \textit{conservative} solution to overestimation bias in RL, while \textit{SQT} shows a clear performance advantage on a wide margin over DDPG, TD3, and TD7 on all tasks.
\end{abstract}

\section{Introduction}

    Reinforcement learning (RL) is the problem of finding an optimal policy, $\pi: \mathcal{S} \xrightarrow[]{} \mathcal{A}$, mapping states to actions, by an agent which makes a decision in an environment and learn by trial and error. We model the problem by an MDP (Markov decision process), let $a_t \in \mathcal{A}$ be an action chosen at timestep $t$ at state $s_t \in \mathcal{S}$, leads to the state $s_{t + 1} \in \mathcal{S}$ in the probability of $P(s_{t + 1} | s_t, a)$ and resulting in the immediate reward of $r_{t + 1}$.

    $Q$-learning \cite{ql} (\ref{ql_algorithm}) is a popular tabular model-free algorithm that suffers from the problem of \textit{overestimation} bias, i.e., its $Q$-values w.r.t. a policy $\pi$ are overestimating the real $Q$-values w.r.t. the policy $\pi$ because it optimizes $Q$-values w.r.t. the "argmax"-policy, which is a biased estimator w.r.t. the expected behavior. This is problematic because those overestimated $Q$-values are propagated further into the whole table through the update process, which finally leads to poor performance.

    Double $Q$-learning \cite{dql} solves this problem by introducing the "double estimator", updating $Q^A$ by $Q^B$ $Q$-values with the action of $a^* = \argmax_{a \in \mathcal{A}} Q^A$, ensures that $a^*$ is unbiased w.r.t. $Q^B$, turning the problem into an \textit{underestimation} bias. Weighted double $Q$-learning \cite{wql} aims to balance between them by weight combining the "single-estimator" with the "double-estimator" in a single update rule.

    In this paper, we introduce a different approach to tackle the problem of \textit{overestimation} bias with a minimal coding effort, using a $Q$-networks disagreement that serves as a penalty for uncertainty.

\section{Background}

    Double $Q$-learning \cite{dql}, a commonly used off-policy algorithm, uses the "double estimator" of the greedy policy, $\mu^A(s) = \argmax_a Q^B(s, a)$. \textit{SQT} uses function approximators parameterized by $\theta^Q$, which it optimizes by minimizing the loss:

    \begin{equation}
        \label{critic}
        L(\theta^Q) = \mathbb{E}_{s_t \sim \rho^{\beta}, a_t \sim \beta, r_t \sim E} [(\underbrace{Q_i(s_t, a_t | \theta^Q) - y_t}_{\text{TD-error.}})^2]
    \end{equation}

    Where:

    \begin{equation}
        \label{sqt_target}
        \begin{split}
            y_t = r(s_t, a_t) + \gamma \underbrace{\mathcal{Q}}_{Q-\text{target formula.}} [Q] (s_{t + 1}, \mu(s_{t+1}) | \theta^Q) - & \\
            \alpha \cdot \underbrace{\textit{SQT}[\mathcal{B}]}_{Q-\text{networks disagreement.}}
        \end{split}
    \end{equation}
    
    While $y_t$ is also dependent on $\theta^Q$, $\mathcal{Q}$ is an ensemble-based $Q$-values operator, $\alpha \in [0, 1]$ is a penalty-parameter, and, $\textit{SQT}$ is a per-batch operator, which is, the mean-batch, std $Q$-networks:

    \begin{equation}
        \textit{SQT}[\mathcal{B}] = \underbrace{\text{mean}_{s \in \mathcal{B}}}_{\text{mean batch.}}[\underbrace{\text{std}_{i=1...N} Q_i(s, a)}_{\text{std } Q-\text{networks.}}], \quad \mathcal{B} \sim \mathcal{D}
    \end{equation}

    While $\mathcal{B}$ is a sampled batch.

    \begin{figure}[H]
        \centering
        \includegraphics[scale=0.35]{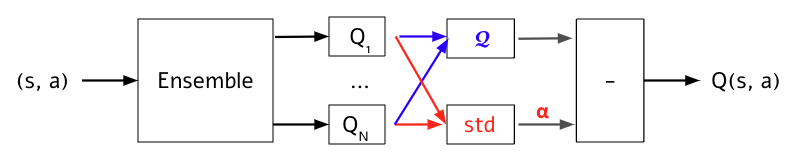}
        \caption{\textit{SQT}'s architecture.}
    \end{figure}

    \subsection{\textit{Overestimation} bias}

        In $Q$-learning with discrete actions, the value estimate is updated with a greedy target $y = r + \gamma \underbrace{\max_a}_{\text{greedy policy.}} Q(s, a)$, however, if the target is susceptible to error, then the maximum over the value along with its error will generally be greater than the true maximum:

        \[
            \underbrace{\mathbb{E}_{\epsilon}[\max_a (Q(s, a)) + \epsilon]}_{\pi_{\text{greedy}} \text{ expectation.}} \geq \underbrace{\max_a Q(s, a)}_{\pi_{\text{greedy}} \text{ real.}}
        \]
        
        As a result, even initially zero-mean error can cause value updates to result in a consistent overestimation bias, which is then propagated through the Bellman equation. This is problematic as errors induced by function approximation are unavoidable.

    \subsection{\textit{Underestimation} bias}

        TD3 \cite{td3} uses a lower bound approximation to the critic. However, relying on this lower bound for exploration is inefficient. By greedily maximizing the lower bound, the policy becomes very concentrated near a maximum. When the critic is inaccurate and the maximum is spurious, this can cause the algorithm to reach suboptimal performance. If the target is susceptible to error, then the minimum over the $Q$-networks of the maximum over the value along with its error will generally be smaller than the true minimum over the $Q$-networks of the maximum:

        \begin{equation}
            \underbrace{\mathbb{E}_{\epsilon}[\underbrace{\min_{i=1...N}}_{\text{lower bound.}} \max_a (Q(s, a)) - \epsilon]}_{\text{expected lower bound.}} \leq \underbrace{\underbrace{\min_{i=1...N}}_{\text{lower bound.}} \max_a Q(s, a)}_{\text{real lower bound.}}
        \end{equation}

\section{\textit{Std} $Q$-target}

    It is easy and straightforward to apply our algorithm, \textit{SQT}, on any Ensemble-based Actor-Critic algorithm with a clear code, such as TD3 and TD7 \cite{td7}, you just need to add a few lines of code into the $Q$-target formula, estimating the \textit{SQT} values and reduce it from the $Q$-target values.

    Since both DDPG, TD3, and TD7, based on DPG \cite{dpg}, \textit{SQT}'s Actor is updated by the mean-batch of the mean $Q$-networks, using the formula:

    \begin{equation}
        \label{actor}
        \nabla_{\theta^{\mu}} J \approx \mathbb{E}_{s_t \sim \rho^{\beta}} [\nabla_{\theta^{\mu}} \underbrace{N^{-1} \sum_{i=1...N} Q_i(s, a | \theta^Q)}_{\text{mean } Q-\text{networks.}} |_{s = s_t, a = \mu(s_t|\theta^{\mu})}]
    \end{equation}

    \textit{SQT}'s contribution is to tackle the problem of \textit{overestimation} bias, which is common in RL algorithms such as DDPG.

    One challenge when using neural networks for RL is that most optimization algorithms assume that the samples are independently and identically distributed. This assumption no longer holds when the samples are generated from exploring sequentially in an environment. Additionally, to efficiently use hardware optimizations, it is essential to learn in minibatches, rather than online. We use a replay buffer to address these issues.

    The replay buffer is a finite-sized cache $\mathcal{R}$. Transitions were sampled from the environment according to the exploration policy and the tuple $(s_t, a_t, r_t, s_{t+1})$ was stored in the replay buffer. When the replay buffer was full the oldest samples were discarded. At each timestep, the actor and critic are updated by sampling a minibatch according to the algorithm's sampling rule.

    As in DDPG, directly implementing $Q$-learning with neural networks proved unstable in many environments. Since the network $Q(s, a | \theta^Q)$ being updated is also used in calculating the target value, the $Q$-update is prone to divergence. As in TD7, we use a target network, using interval-based target updates. We create a copy of the actor and critic networks, $Q'(s, a | \theta^{Q'})$ and $\mu' (s|\theta^{\mu'})$ respectively, that are used for calculating the target values. The weights of these target networks are then updated by having them slowly track the learned networks:

    \begin{equation}
        \label{target}
        \theta' \xleftarrow[]{} \underbrace{\theta}_{\text{new parameters.}}
    \end{equation}

    At a specified timesteps interval $t$.

    A major challenge of learning in continuous action spaces is exploration. As in DDPG, we constructed an exploration policy $\mu$ by adding noise sampled from a noise process $\mathcal{N}$ to our actor policy:

    \begin{equation}
        \label{action}
        \mu'(s_t) = \mu(s_t|\theta^{\mu}_t) + \underbrace{\mathcal{N}}_{\text{random noise}.}
    \end{equation}
    
    $\mathcal{N}$ can suit the environment. 

    To \underline{summarize}, the \textit{SQT} algorithm can be summarized into a single line: \textbf{pessimistic $Q$-formula} reduces a $Q$-function disagreement term from the algorithm's $Q$-values.

    \textit{SQT} algorithm can be summarized by the following pseudo-code:

    \subsection{Algorithm}

        \begin{algorithm}[H]
            \begin{algorithmic}[1]            
                \FOR{each iteration $t$}     
                              
                    \STATE Take a step $a$ in state $s$ by \ref{action}.
                    \STATE Store tuple $\mathcal{D} \xleftarrow[]{} \mathcal{D} \cup \{(s, a, r, s', d)\}$.            
                    \FOR{each iteration $g \in G$}
                    
                        \STATE Sample batch, $\mathcal{B} \sim \mathcal{D}$, by the algorithm's sampling-rule.                                
                        \STATE Compute $Q$-target $y$, by \ref{sqt_target}.

                        \STATE Update critic parameters by \ref{critic}.
                        \STATE Update actor parameters by \ref{actor}.
                        \STATE Update target parameters by \ref{target} at the specified interval.
            
                    \ENDFOR                
                \ENDFOR
            \end{algorithmic}            
            \caption{\textit{SQT}}
        
    \end{algorithm}

\section{Experiments}

    We constructed simulated physical environments of varying levels of difficulty to test our algorithm. This included locomotion tasks such as humanoid, walker, ant, cheetah, swimmer, and hopper. In all domains, the actions were torques applied to the actuated joints. These environments were simulated using MuJoCo \cite{mujoco} and Bullet \cite{bullet}.

    As we thought, the more complex environment including humanoid, walker, ant, cheetah, swimmer, and hopper, demonstrates well the clear advantage of \textit{SQT}'s approach over TD3 and TD7 by a significant average seeds performance advantage in all of these problems.

    We implement \textit{SQT} on Actor-Critic settings \footnote{\textit{SQT}'s code: \url{https://github.com/anonymouszxcv16/SQT}}, while the $Q$-target formula is TD3/TD7 $Q$-target formula, and test it on MuJoCo/Bullet \cite{mujoco}, \cite{bullet} benchmark, by the six popular environments, which are:

    \begin{enumerate}
        \item \textbf{Humanoid}: a locomotion-based humanoid with 2 legs and 2-arms with the target of walking forward without falling over.
        \item \textbf{Walker2d}: a locomotion-based walker with 2 legs with the target of walking forward without losing stability.
        \item \textbf{Ant}: a locomotion-based ant with 4 legs with the target of moving forward without turnover.
        \item \textbf{HalfCheetah}: a locomotion-based cheetah with 4 legs with the target of moving forward as fast as possible.
        \item \textbf{Swimmer}: a locomotion-based swimmer with 3 segments with the target of moving as fast as possible toward the right.
        \item \textbf{Hopper}: a locomotion-based hopper with a 1-legged figure that consists of four main body parts with the goal is to making hops that move in forward.
    \end{enumerate}

    We also run DDPG, TD3, and TD7 on the same machine with the same compute resources with a total of $5$ seeds: $\{0...4\}$, reporting the average rewards of the maximum performance snapshot, comparing \textit{SQT} to the competitor in percentages and summarizes the total improvement, and plot the mean seeds average rewards.
    
    The following table summarizes the results of \textit{SQT} vs. TD7 on MuJoCo tasks:

    \begin{table}[H]
        \begin{center}
            
            \begin{tabular}{ l | c c c}
             Environment & TD7 & \textit{SQT} & Improvement \\
             \hline
                Humanoid-v2 & 6,783.7 & \textbf{8,144.7} & +\textbf{20.1}\% \\
                Walker2d-v2 & 6,058.9 & \textbf{7,121.8} & +\textbf{17.5}\% \\
                Ant-v2 & 8,300.6 & \textbf{8,906.2} & +\textbf{7.3}\% \\
            \hline    
            & & & \textbf{+44.9\%}\\
            \end{tabular}
            \caption{\textit{SQT} vs TD7.}
        \end{center}    
    \end{table}    

    \textit{SQT} shows a clear performance superiority over TD7 on all the benchmarks, especially on humanoid and walker. We hypothesize that the reason is that humanoid and walker can lose stability or fall over on taking a suboptimal action which can lead to poor performance and finally prevent the algorithm from converging on some seeds on TD7, while on \textit{SQT}, due to its conservative nature, it converges to optimum on all seeds.
    
    The following table summarizes the results of \textit{SQT} vs. TD3 on MuJoCo tasks:
    
    \begin{table}[H]
        \begin{center}
            
            \begin{tabular}{ l | c c c}
             Environment & TD3 & \textit{SQT} & Improvement \\
             
             \hline    
                Humanoid-v2 & 5,043.4 & \textbf{6,648.3} & \textbf{+31.8\%} \\
                Walker2d-v2 & \textbf{5,459.2} & 5,458.3 & -0.0\% \\
                Ant-v2 & 6,432.5 & \textbf{6,707.2} & \textbf{+4.3\%} \\    
            \hline            
            
            & & & \textbf{+36.1\%}\\            
            
            \end{tabular}
            \caption{\textit{SQT} vs. TD3.}
        \end{center}
    \end{table}

    \textit{SQT} shows a clear performance superiority over TD3 on all the benchmarks, especially on humanoid. We hypothesize that the reason is that humanoid can fall over on taking a suboptimal action which can lead to poor performance and finally prevent the algorithm from converging on some seeds on TD3, while on \textit{SQT}, due to its conservative nature, it converges to optimum on all seeds.
    
    The following plot summarizes the results of \textit{SQT} when applied on top of TD7 vs. DDPG, TD3, and TD7 on the ant bullet task:

    \begin{figure}[H]
        \centering
        \includegraphics[scale=0.5]{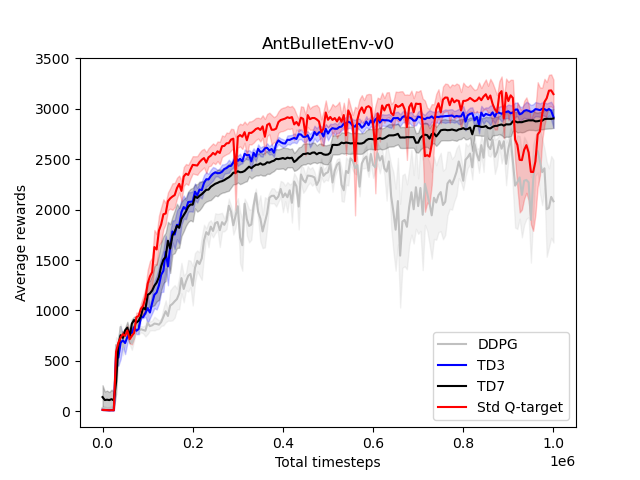}
        \caption{\textit{SQT} when applied on top of TD7 vs. DDPG, TD3 and TD7 on ant bullet.}
    \end{figure}

    \textit{SQT} when applied on top of TD7 shows a wide-margin performance advantage over all competitors, especially over TD7 whose performance falls under TD3. We hypothesize that the reason is a scenario when the ant turns over due to a suboptimal action, which finally leads to poor performance on TD7, a scenario which does not occur in \textit{SQT}, due to its conservative nature.

    DDPG exhibits a noticeable performance drawback compared to its counterparts. We posit that this disadvantage is particularly pronounced in risky environments, such as those involving ants that may overturn when the agent undertakes a risky action, resulting in a substantial point loss. This is attributed to DDPG's inherent problem of \textit{overestimation} bias, wherein it tends to overvalue suboptimal state-action pairs. Consequently, this \textit{overestimation} leads to the selection of risky actions, culminating in poor performance. For instance, an ant might overturn or make unfavorable decisions during exploration, adversely affecting overall performance.
    
    The following plot summarizes the results of \textit{SQT} when applied on top of TD7 vs. DDPG, TD3, and TD7 on the cheetah bullet task:

    \begin{figure}[H]
        \centering
        \includegraphics[scale=0.5]{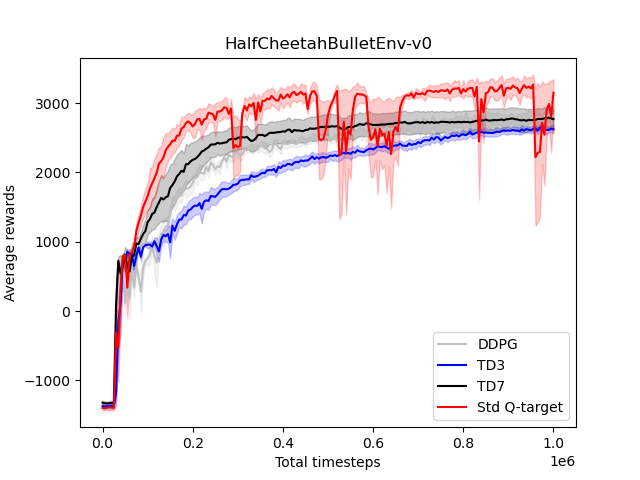}
        \caption{\textit{SQT} when applied on top of TD7 vs. DDPG, TD3 and TD7 on cheetah bullet.}
    \end{figure}    

    \textit{SQT} when applied on top of TD7 shows a wide-margin performance advantage over all competitors. We hypothesize that the reason is a scenario when the cheetah performs a risky explorative action, i.e. a sharp movement, causing poor data generation and finally a suboptimal performance, a scenario which does not occur in \textit{SQT}, due to its conservative nature.

    The following plot summarizes the results of \textit{SQT} when applied on top of TD7 vs. DDPG, TD3, and TD7 on the swimmer task:

    \begin{figure}[H]
        \centering
        \includegraphics[scale=0.5]{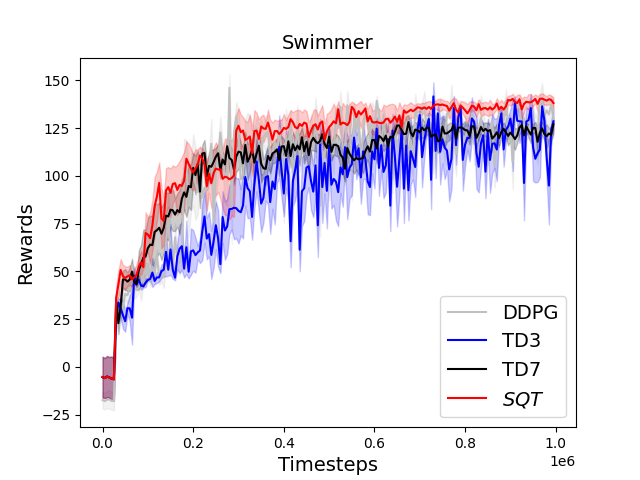}
        \caption{\textit{SQT} when applied on top of TD7 vs. DDPG, TD3 and TD7 on swimmer.}
    \end{figure}    

    \textit{SQT} when applied on top of TD7 shows a clear performance advantage over all competitors when keeping stable results per-seed. Our hypothesis to \textit{SQT}'s advantage on the swimmer, is due to the swimmer's complexity which requires a safe conservative critic's updates, and punishes for sharp mistaken action, which is a great fit to \textit{SQT} that uses a conservative critic's updates. TD3, due to its less conservative updates than \textit{SQT}'s, can choose poor actions sometimes for exploration, while \textit{SQT} is more sensitive to poor choices, due to its conservative nature.

    The following plot summarizes the results of \textit{SQT} when applied on top of TD7 vs. DDPG, TD3, and TD7 on the hopper-bullet task:

   \begin{figure}[H]
        \centering
        \includegraphics[scale=0.5]{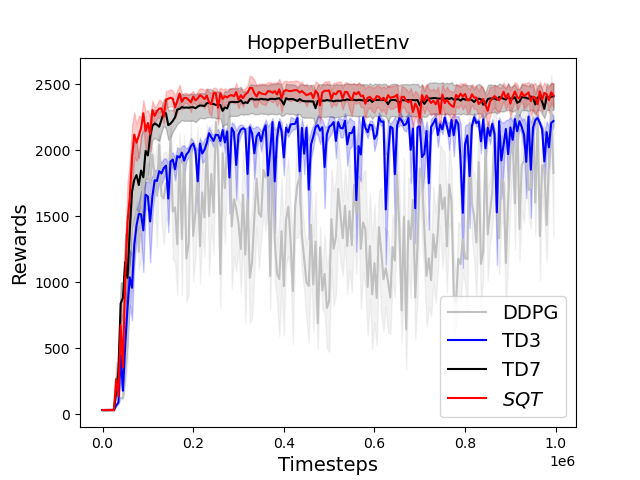}
        \caption{\textit{SQT} when applied on top of TD7 vs. DDPG, TD3 and TD7 on hopper-bullet.}
    \end{figure}    

    \textit{SQT} when applied on top of TD7 shows a clear performance advantage over all competitors when keeping stable results per-seed. We hypothesize that the reason is a scenario when the hopper performs a risky explorative action, i.e. poor movement, causing the hopper to lose stability and lose performance, a scenario which does not occur in \textit{SQT}, due to its conservative nature.

    TD3 displays a significant level of performance volatility in contrast to \textit{SQT}. Our conjecture centers around situations where TD3 opts for a risky action, one with a high probability of resulting in poor performance in favor of exploration. This tendency leads to inconsistent results across different seed values, a scenario that does not occur in \textit{SQT}, as it prioritizes safety over exploration.

    In the case of DDPG, there is an evident performance drawback when compared to other models. We theorize that this limitation becomes particularly apparent in environments with instability, such as those involving a hopper that may lose stability when the agent takes a risky action, resulting in a significant loss of points. This is linked to DDPG's inherent issue of \textit{overestimation} bias, where it tends to overvalue suboptimal state-action pairs. As a consequence, this tendency to \textit{overestimate} leads to the selection of risky actions, resulting in subpar performance. For example, a hopper might experience instability or make unfavorable decisions during exploration, negatively impacting overall performance.
    
\section{Related work}

    $Q$-learning's algorithm can be summarized with the following pseudo-code:

    \label{ql_algorithm}
    \begin{algorithm}[H]
        \begin{algorithmic}[1]            
            \FOR{each iteration}
                \STATE Observe its current state $s_t$.
                \STATE Selects and acts $a_t$.
                \STATE Observes the subsequent state $s'_t$.
                \STATE Receives an immediate payoff $r_t$.
                \STATE Adjust its $Q_{t - 1}$ values using a learning factor $\alpha_t$, according to:
                    \begin{equation}
                        \begin{split}
                            Q_t(s_t, a_t) = (1 - \alpha_t) Q_{t - 1} (s_t, a_t) + & \\
                            \alpha_t [r_t + \gamma \underbrace{V_{t - 1} (s'_t)}_{\max_{a'} \{ Q_{t - 1} (s', a') \}}]
                        \end{split}
                    \end{equation}
            \ENDFOR
        \end{algorithmic}            
    \caption{$Q$-learning}        
    \end{algorithm}

    \textit{Conservative} policy creation (CPI) \cite{cpi} is an algorithm that finds an approximately optimal policy given access to a restart distribution and an approximate greedy policy chooser using the following \textit{conservative} policy update rule:

    \begin{equation}
        \pi_{\textit{new}} = (1 - \alpha) \underbrace{\pi (a; s)}_{\pi_{\text{old}}.} + \alpha \pi' (a; s)
    \end{equation}

    Which precedes a cure for a blow of the \textit{overestimation} bias problem.

    TD3 \cite{td3} is a proven minimalistic solution to DDPG's \cite{ddpg} overestimation bias, it just adds another $Q$-function, inspired by Double $Q$-learning's "double estimator" \cite{dql}, and takes the minimum $Q$-network $Q$-values as a \textit{conservative} estimator. Although TD3 has been demonstrated as promising, \textbf{it solves a problem by introducing another problem} which is \textit{underestimation} bias, resulting in a poor exploration and finally in suboptimal performance.

    TD7 \cite{td7} is a simple actor-critic model-free algorithm based on TD3 that can be summarized to one single main component which is SALE (state-action learned-embedding), i.e., TD7 learns an encoding of the transition dynamics of the MDP, $T: \mathcal{S} \times \mathcal{A} \xrightarrow[]{} \mathcal{S}$, mapping state-action tuple into the state, in the form of state-encoding, $z^s$, and a state-action encoding, $z^{sa}$, which uses as an input to the $Q$-function, and to the policy $\pi$, along with the state-action tuple. TD7 demonstrated a clear performance advantage over TD3 when tested on MuJoCo tasks.

    Batch-constrained deep $Q$-learning (BCQ) \cite{bcq} is a \textit{conservative} RL algorithm that combines the $\min$ $Q$-networks, with the, $\max$ $Q$-networks, resulting in a weighted $Q$-target which favors the minimum, as a modest solution to the problem of \textit{overestimation} bias of DDPG.

    BCQ $Q$-target formula is:

    \begin{equation}
        y = r + \gamma \max_{a} [\underbrace{\lambda \min_{j=1, 2} Q_j (s', a) + (1 - \lambda) \max_{j=1, 2} Q_j (s, a)}_{\text{MaxMin combination}.}]
    \end{equation}

    MaxMin $Q$-learning \cite{maxmin_ql} is an extension of TD3 that introduces a solution to \textit{overestimation} bias problem on RL algorithms, by accommodating N $Q$-functions, allowing flexible control over bias levels—whether it leans toward \textit{overestimation} or \textit{underestimation}—determined by the chosen value of $N$. 

    This contribution holds significant value in two aspects: Firstly, in its generic extension of TD3 to handle $N$ $Q$-functions (contrasting with TD3's limitation to two functions). Secondly, it lays out the formal theoretical properties of the MaxMin algorithm, offering a solid theoretical foundation for subsequent algorithms. Specifically, it proves convergence of MaxMin $Q$-learning in tabular settings to the optimal $Q$-values ($Q^*$) and establishes fundamental principles to verify convergence in ensemble RL algorithms, offering a generic framework for broader algorithmic development.

    In MaxMin, the $Q$-values are estimated using this formula:

    \[
        y = r + \gamma \underbrace{\min_{i=1...N} Q_i (s', \pi_{\phi}(s'))}_{\text{$Q$-networks agreement.}}
    \]
    
    Which estimates the $Q$-values by selecting the minimum among a set of $Q$-networks for the given state-action pair.

    Model-based policy optimization (MBPO) \cite{mbpo} is a simple procedure of using short model-generated rollouts branched from real data, which, surpasses the sample efficiency of prior model-based methods, and matches the asymptotic performance of the best model-free algorithms.

    MBPO uses a branched rollout to collect data: starting from the previous policy's state distribution, $d_{\pi_{D}} (s)$, and then taking $k$ steps with the new policy, $p_{\theta}$, inspired by Dyna \cite{dyna} (which do a branched rollout with $k = 1$).

    Randomized ensembled double $Q$-learning (REDQ) \cite{redq} is a novel deep ensemble RL algorithm, that, formulates a novel $Q$-values formula, and, uses an Update-To-Data (UTD) ratio $G \gg$ 1.
    
    It is the first model-free algorithm to use a UTD ratio $\gg 1$, reaching a competitive sample-complexity performance to MBPO \cite{mbpo} without a model. REDQ's $Q$-target formula is:

    \begin{equation}
        y = r + \gamma (\min_{i \in \mathcal{M}} Q_i(s', a') - \alpha \underbrace{\log \pi(a' | s')}_{\text{confidence penalty}.}), \quad \mathcal{M} \sim \{1...N\}
    \end{equation}

    The $\beta$-\textit{pessimistic} $Q$-learning \cite{beta_pessimistic}, strikes a balance between the extreme \textit{optimism} seen in standard $Q$-learning and the extreme \textit{pessimism} of \textit{MiniMax} $Q$-learning \cite{minimax_ql} (\ref{minimax_algorithm}). This approach ensures the robustness of safe areas within the state space regarding the chosen action. In this context, the $\beta$-\textit{pessimistic} action-values estimate the anticipated value of taking an action, followed by actions that have the highest value with a probability of $1 - \beta$, or the lowest value with a probability of $\beta$. When $\beta = 0$, it mirrors standard $Q$-learning, and when $\beta = 1$, it aligns with the principles of MiniMax $Q$-learning \cite{minimax_ql}.

    The $Q$-value estimates in $\beta$-\textit{pessimistic} $Q$-learning are calculated using the following formula:

    \[
    \begin{split}
        y = r + \gamma \bigg[ (1 - \beta) \cdot \max_{a \in \mathcal{A}} Q(s, a) + \beta \cdot \min_{a \in \mathcal{A}} Q(s, a) \bigg] = \\
        r + \gamma \Bigg[ \max_{a \in \mathcal{A}} Q(s, a) - \beta \cdot \bigg( \underbrace{\max_{a \in \mathcal{A}} Q(s, a) - \min_{a \in \mathcal{A}} Q(s, a)}_{\text{MaxMin action gap}.} \bigg) \Bigg]
    \end{split}
    \]
    
    This equation combines the reward ($r$), discounted future rewards ($\gamma$), and a weighted combination of the maximum and minimum $Q$-values, where the weight $\beta$ modifies the variance between the maximum and minimum action values.

    $Q$-$\kappa$ \cite{q_kappa} is a \textit{conservative} tabular $Q$-learning algorithm based on a single $Q$-based operator, the robust TD-operator $\kappa$:

    \[
        \begin{split}
            \delta_t = r_{t + 1} + \gamma [(1 - \kappa) \underbrace{\max_a Q(s_{t + 1}, a)}_{\text{agent's control.}} + \kappa \underbrace{\min_a Q(s_{t + 1}, a)}_{\text{adversarial's control.}}] - & \\
            Q(s_t, a_t)
        \end{split}
    \]

     Suppose a $Q$-learning agent must learn a robust policy against a malicious adversary who could, take over control in the next state, $s_{t + 1}$. The value of the next state, $s_{t + 1}$, thus depends on who is in control: if the agent is in control, she can choose an optimal action that maximizes expected return; or if the adversary is in control he might aim to minimize the expected return.

    MiniMax $Q$-learning algorithm is a $Q$-learning-based tabular algorithm that optimizes the $Q$-function w.r.t. a risk-based criterion of the worst-case discounted average rewards. MiniMax $Q$-learning's \cite{minimax_ql} algorithm can be summarized with the following pseudo-code:

    \label{minimax_algorithm}
    \begin{algorithm}[H]
        \begin{algorithmic}[1]            
            \FOR{each iteration}
                \STATE $s :=$ starting state of the current episode.
                \STATE Select an action $a \in \mathcal{A}$ and execute it.
                \STATE $s' :=$ successor state of current episode.
                \STATE $r :=$ immediate cost of current episode.
                \STATE $Q(s, a) := \max [\underbrace{Q(s, a)}_{\text{old.}}, \underbrace{r + \gamma \underbrace{\min_{a' \in \mathcal{A}} Q(s', a')}_{\text{pessimistic.}}}_{Q-\text{target}}]$
            \ENDFOR
        \end{algorithmic}            
    \caption{MiniMax $Q$-learning}        
    \end{algorithm}

\section{Conclusion}

    \textit{SQT} is a novel $Q$-learning-based, model-free, online, on-policy, Actor-Critic, \textit{pessimistic} algorithm, tackles the problem of \textit{overestimation} bias of RL algorithms such as DDPG, by using a single, core, simple idea of adding an "uncertainty penalty" which based on a $Q$-networks disagreement, into the $Q$-target formula.

    We implement \textit{SQT} on top of TD3/TD7, as an ensemble-based Actor-Critic algorithm, and test it against DDPG, TD3, and TD7, as the state-of-the-art (SOTA) Actor-Critic, model-free, online algorithms, on seven popular MuJoCo and Bullet locomotion tasks, which are: humanoid, walker, ant, swimmer, ant-bullet, cheetah-bullet, and hopper-bullet. Our results show a clear performance advantage to \textit{SQT}, on a wide margin, in all the tested tasks over DDPG, TD3, and TD7. Demonstrating the superiority of \textit{SQT}'s \textit{conservative} $Q$-values formula over TD3's $Q$-values formula on those tasks. 

    In conclusion, our final statement asserts that addressing the issue of \textit{overestimation} bias in a model-free, online, on-policy, Actor-Critic, ensemble-based algorithm can be achieved by simply diminishing the \textit{SQT} term in the $Q$-target formula. This reduction acts as an "uncertainty penalty," yielding a \textit{conservative} formulation for $Q$-values.
    
\bibliographystyle{named}
\bibliography{ijcai24}

\end{document}